\documentclass[10pt,conference,a4paper,twocolumn] {IEEEtran}

\usepackage{graphicx}
\usepackage{epic}
\usepackage{figsize}
\usepackage{float}
\usepackage{hyperref}
\usepackage[section]{placeins}
\usepackage{cite}
\usepackage{amsmath}
\usepackage{amssymb}
\usepackage{amsfonts}

\newcommand{\email}[1]{{\fontfamily{cmtt}\selectfont#1\normalfont}}

\IEEEoverridecommandlockouts
\usepackage{cite}
\usepackage{amsmath,amssymb,amsfonts}
\usepackage{algorithmic}
\usepackage{graphicx}
\usepackage{textcomp}
\usepackage{xcolor}
\def\BibTeX{{\rm B\kern-.05em{\sc i\kern-.025em b}\kern-.08em
    T\kern-.1667em\lower.7ex\hbox{E}\kern-.125emX}}

\title{A Twitter-Driven Deep Learning Mechanism for the Determination of Vehicle Hijacking Spots in Cities}

\author{\IEEEauthorblockN{Taahir Aiyoob Patel, Clement N. Nyirenda}
\IEEEauthorblockA{\emph{Department of Computer Science,}\\
\emph{Faculty of Natural Sciences,}\\
\emph{The University of the Western Cape}\\
\emph{Bellville, Cape Town 7535} \\
\emph{South Africa} \\
$^1$\email{3759655@myuwc.ac.za}\\
$^2$\email{cnyirenda@uwc.ac.za}
}}

\begin{document}

\maketitle

\begin{abstract}
Vehicle hijacking is one of the leading crimes in many cities. For instance, in South Africa, drivers must constantly remain vigilant on the road in order to ensure that they do not become hijacking victims. This work is aimed at developing a map depicting hijacking spots in a city by using Twitter data. Tweets, which include the keyword “hijacking”, are obtained in a designated city of Cape Town, in this work. In order to extract relevant tweets, these tweets are analyzed by using the following machine learning techniques: 1) a Multi-layer Feed-forward Neural Network (MLFNN); 2) Convolutional Neural Network; and Bidirectional Encoder Representations from Transformers (BERT). Through training and testing, CNN achieved an accuracy of 99.66\%, while MLFNN and BERT achieve accuracies of 98.99\% and 73.99\% respectively. In terms of Recall, Precision and F1-score, CNN also achieved the best results. Therefore, CNN was used for the identification of relevant tweets. The relevant reports that it generates are visually presented on a points map of the City of Cape Town. This work used a small dataset of 426 tweets. In future, the use of evolutionary computation will be explored for purposes of optimizing the deep learning models. A mobile application is under development to make this information usable by the general public.
\end{abstract}


\section{Introduction}
Drivers are constantly at risk of being hijacked. Locals may be able to avoid certain areas due to personal knowledge; it is, however, not uncommon to fall victim nonetheless. As the number of drivers increases, the risk becomes greater. Between July and September 2021, in South Africa there was a total of 4973 hijackings \cite{b3}. This shows the magnitude of the issue. 

Social media platforms such as twitter provide a constant flow of information. Twitter, itself, is commonly known for allowing users to discuss their daily thoughts and events. It is also used by news outlets to inform users of current real-time events. This inspires the research of using Twitter as a real-time data source for car hijacking incident reports. Obtaining hijacking crime data from Law enforcement agencies can be a difficult task, for reasons such as there not being updated collections of data, or perhaps difficulty in obtaining administrative approval. Generally, the supplied reports and analytical data are usually presented in large time increments, such as monthly, annually, etc. Therefore, with the use of Twitter as the main source of data obtainment, the data can be used and updated in real-time. However, an issue arises from using social media as the data source, and that is the issue of relevancy. Due to the lack of restrictions on tweets, a large amount of the obtained tweets could be irrelevant to the topic. For example, searching for tweets containing the keyword of hijacking, would return incident report tweets, as well as tweets discussing a different topic, however containing the keyword. Therefore, there is a need for machine learning to separate the off-topic irrelevant tweets, from the relevant incident reports.

Piña‐García and Ramírez‐Ramírez \cite{b4} proposed using a linear predictive model to evaluate the performance of google trends in predicting crime rates based on a weekly analysis. This was done by presenting multiple sections in the report which addressed different factors influencing the proposal such as a section presenting heatmaps of the number of crimes in Mexico City by per borough, then by neighborhoods \cite{b4}. They were able to prove that Google trends and Twitter can be used to generate estimates of crime occurrences. They presented this by using a linear predictive model. This proposal introduced the possibility of using Google trends and Twitter to track crime trends and
analysing it further to gain relevant information.

Mata \textit{et al} \cite{b5} developed an application to help locate safe routes of travel using estimations defined from crime rates. They demonstrated their proposal by providing the algorithms and pseudo code they used, as well as examples of their proposed application user interface performing the proposed functionalities. Through their research, they were able to obtain the safest traveling route
by using the Bayes algorithm with a certainty degree of around 75\% effectiveness \cite{b5}. Their research provides further insight for the research done in this paper by using Twitter to gain the reports.

Lal \textit{et al}  \cite{b6} obtained and analysed a total of 369 tweets into classes of Crime and Not-Crime respectively. To conduct their study they used a text minding approach whereby they employed 4 main classifiers for testing, namely Naive Bayesian (NB), Random Forest (RF), J48 and ZeroR. The RF classifier performed the best with an accuracy value of 98.1\%, but it had the longest model building time of 7.24s. On the other hand, the ZeroR classifier was the worst performing classifier. In terms receiver operating characteristic (ROC) analysis, NB, J48, and RF gave gave the best performance \cite{b6}. 

The work in \cite{b6} inspired the work proposed in this paper. While Twitter has been used for generic crime prediction, so far no work has been reported on Twitter-based determination of vehicle hijacking spots. Since Twitter contains both relevant and irrelevant or fake tweets, the need for the determination of relevant tweets by performing classification of “hijacking” and “non-hijacking” arises naturally. Therefore, this paper addresses this issue by using a similarly small datasets to the one that was used in \cite{b6}. While the work in \cite{b6} used shallow learning techniques, this work uses three deep learning techniques, namely: a Convolutional Neural Network (CNN), an Multi-layer Feed Forward Neural Network (MLFNN), and a Bidirectional Encoder Representations from Transformers (BERT) model. These models are trained by using 296 tweets and tested by using 130 tweets. The best performing model is adopted and used to identify relevant hijacking reports from the tweets database, and the vehicle hijacking locations are extracted from the tweets to indicate them on the visual maps of hijacking spots. In this study, the City of Cape Town is used. But the approach can be used in any city as long as the tweets are in English. 

This paper is organised as follows. A brief overview into the Machine Learning mechanisms used in this work is presented in Section II. Section III presents the materials and methods we employed in this work. The results are presented in Section IV. Lastly, the paper is concluded in Section V. 

\section{A Brief Overview on Deep Learning Mechanisms Employed in This Work}
This section presents a brief overview on the three machine learning algorithms used in this work. 

\subsection{Multi-Layer Feed-Forward Neural Network}
A Multi-Layer Feed-Forward Neural Network \cite{b7} is a neural network whereby each layer is fully connected to each other to the subsequent layer. It consists of an input layer, output layer, as well as a number of hidden layers. A MLFNN is generally used for supervised learning, which implies that the model is trained using a dataset of inputs and their correct outputs. When training a MLFNN, the model is given a set of $N$ input-output($x$,$y$) pairs $X$. The training process involves the constant iterations of updating values of the weights of the perceptrons $j$ in layers $l_k$ for the incoming node $i$, as well as the bias $b^k_i$ for the output $o_i$ on the input $x_i$, with the aim of minimizing the mean squared error \cite{b8}, which is given by 
\begin{equation}
E(X)=\frac{1}{2N} \sum_{i=1}^{N} (o_i-y_i)^2\,
\end{equation}

\subsection{Convolutional Neural Network}
Convolutional Neural Network(CNN) \cite{b9} is generally comprised of 3 types of layers, namely Convolutional layers, Max Pooling layers, and lastly the fully-connected layers. The Convolutional layer is aimed at outputting a high value for a given position if there is a convolutional feature in the given position, otherwise, a low value is outputted \cite{b10}. This value is obtained by using
\begin{equation}
h_{i,j}=\sum_{k=1}^{m}\sum_{l=1}^{m}  w_{k,l} {x_{i+k-1,j+l-1}}\,
\end{equation}
where $m$ is the kernel width and height, $h$ is the convolution output, $X$ is the input, and $W$ is the convolution kernel. The max pooling layer is aimed at outputting the maximum value of the input which falls on the kernel \cite{b10}, which is defined by
\begin{equation}
p_{i,j}= max\{w_{i+k-1,j+l-1}\forall{1\leq k\leq m \:and\: 1 \leq l\leq m}\}\,
\end{equation}

A CNN is generally aimed at image recognition. Due to the nature of CNN for image data, the standard format is Conv2D. However, in this work, a one dimensional CNN, Conv1D is employed because of the textual data that is being used.

\subsection{Bidirectional Encoder Representations from Transformers}
The bidirectional Encoder Representations from Transformers(BERT) model \cite{b11} employs Transformers, which allow it to learn the contextual relationship between words of a sentence. The BERT model was adopted with the aim of implementing Transfer learning. Text classification is a common area of interest for transfer learning, due the advancement of Natural Language Processing (NLP). The BERT model incorporates a multi-layer bidirectional Transformer encoder which comprises of a stack of 6 identical layers, with each having a set of two sub-layers \cite{b12}. The first sub layer is a multi-head self-attention mechanism, whereas the second is a position-wise fully connected feed-forward network. For the multi-head self-attention, the scaled dot-product attention is required, which is defined by
\begin{equation}
Attention(Q,K,V)=softmax(\frac{QK^T}{\sqrt{d_k}})V,
\end{equation}
where $Q$, $K$, $V$, are the matrices of the queries, keys, values respectively, and $d_k$ is the dimension of the matrices, $Q$ and $K$. The multi-heads $head_i$ are computed by using
\begin{equation}
head_i = Attention(QW^Q_i,KW^K_i, VW^V_i) 
\end{equation}
where $QW$,$KW$, and $VW$, are the query, key, and value matrices respectively. From this, the multi-head self-attention is performed by using
\begin{equation}
MultiHead(Q,K,V)=Concat(head_1,....,head_h)W^O
\end{equation}
where $W$ is a weight matrix. 

\section{Materials and Methods}
\begin{figure}[htbp]
    \centerline{\includegraphics[scale=0.6]{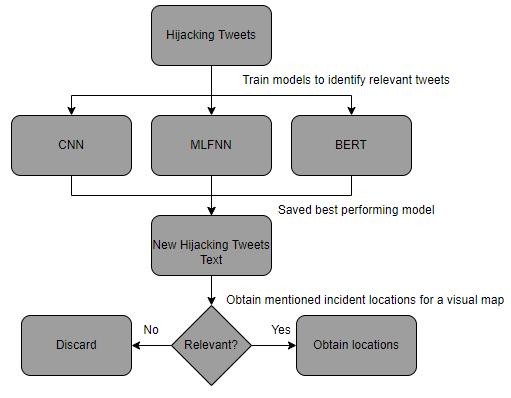}}
    \caption{System flow diagram}
    \label{ReducedFlow}
\end{figure}

This section presents the materials and methods that were used in order to realize the goals of this study. First of all, two sets of datasets had to be generated for training and for testing processes. After that, data was processed in order to align with the deep learning models used in this study. Finally the models were trained and testing was conducted. These approaches are explained in detail in the next subsections. The system flow diagram can be seen in Fig.~\ref{ReducedFlow}.

\subsection{Data Generation and Pre-processing}
This work is concerned with obtaining raw, unedited tweets, and understanding whether they denote genuine hijacking reports, or they are just irrelevant tweets. For purposes of this study, the City of Cape was chosen as the target area. Tweets containing the words "hijacking" and "Cape Town" were collected by using a small python program focusing on tweets in a 50km radius from the center of Cape Town. This was done with the Tweepy \cite{b13} python library. This data was then pushed into a MySQL data for storage. The total number of tweets collected was 426. These tweets were extracted into a into an Excel spreadsheet, where each tweet was then manually indicated as relevant or irrelevant by applying a code of 1 and 0 respectively. There were eventually divided into two sets as follows: 1) 296 tweets constituted the training set(76 relevant, 220 irrelevant); 2) 130 constituted the testing set (29 relevant, 101 irrelevant).

To prepare the data for training, standard pre-processing was conducted whereby leading or trailing spaces were removed followed by converting the text to lower case, and lastly, removing stop words. Inspired by the work of Wang \textit{et al} \cite{b14}, the TfidfVectorizer was also incorporated in the MLFNN and CNN mechanisms. For text classification with Natural Language Processing (NLP), there are multiple different strategies to do this, such as the Bag-of-Words methods like the Term Frequency - Inverse Document Frequency (TF-IDF) method, or through language models such as BERT, or perhaps through Word2Vec, such as the approach of Jang \textit{et al} \cite{b15}. The TF-IDF identifies the number of occurrences of a word in a document, as well as the measure of the rarity of the word in the document. The BERT method incorporates the ELMo context embedding \cite{b16}, as well as multiple Transformers. For BERT, each vector which is assigned to a word, is a function of the sentence as a whole, therefore, based on the context each word has the possibility of having more than one vector \cite{b17}. As a result, the TfidfVectorizer was not required for the BERT implementation.

\subsection{Training and Testing}
TensorFlow \cite{b18} and Python 3.8 were used to train the model. The Adam optimizer was used for optimization, along with the binary cross-entropy loss. A batch size of 32 was used over 20 epochs. A validation split of 0.2 was also incorporated. The training was done on a home desktop with a single AMD RX570 GPU running a Windows 10 Operating System.

The following performance metrics were used:
\begin{itemize}
    \item \textit{Accuracy}: This is a percentage measurement of how well the model predicted a true value. This is done during the training and validation phase.
    \item \textit{Loss}: This is the summation of the number of errors the model makes. The loss function, binary cross-entropy performs a calculation of the difference between the actual value, and predicted values. It then provides a penalty score based on how far the predicted value is from the actual. This was calculated during the training and validation phase.
    \item \textit{Precision}: This is the measurement of how much the model correctly predicted a positive, over the number of positive predictions as a whole.
    \item \textit{Recall}: This is the measurement of how much the model correctly predicted a positive, over all the actual positives.
    \item \textit{F1-score}: This is the weighted averages of the Precision and Recall.
\end{itemize}
Precision, recall, and F1-score are used during testing. This approach was adopted from the work in \cite{b6, b19}.  The Accuracy and F1-Score were also used in the CNN implementation in \cite{b15}. 

Model specific measures for the training process were implemented as follows:
\begin{enumerate}
    \item \textit{Convolutional Neural Network}: For the development of the CNN, the base model consisted of a single Convolutional and Max Pooling Layer. The performance of this model was used as a base, then the number of layers was increased by adding Convolutional and Max Pooling layers. This was done twice, increasing the number of Convolution and MaxPooling layers by 1 each time. Three models were therefore obtained: Single Conv1D/MaxPooling; Double Conv1D/MaxPooling; and Triple Conv1D/MaxPooling. Fig.~\ref{cfig} illustrates the Double Conv1D/Maxpooling architecture. 
    \begin{figure*}[htbp]
    \centerline{\includegraphics[scale=0.75]{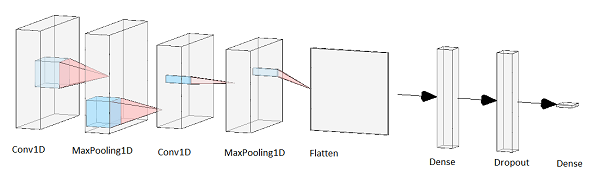}}
    \caption{The Double Conv1D/Maxpooling architecture.}
    \label{cfig}
     \end{figure*}
    \item \textit{Multi-Layer Feed-Forward Neural Network}: Similar to the CNN, a base model was selected and the layout was adjusted to compare performance. The base model was influenced by the work in \cite{b14}. The initial model that was selected comprised of 2 Dense Layers. The number of layers was then increased by 1 each time and the performance was observed before selecting the preferred model. Fig.~\ref{MLFNNfig} illustrates the MLFNN architecture with four dense layers.
    \begin{figure}[htbp]
    \centerline{\includegraphics[scale=0.7]{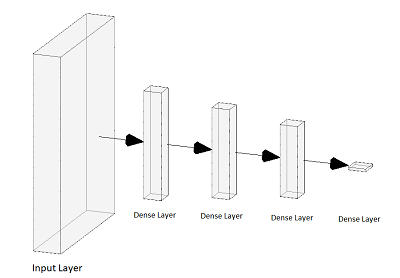}}
    \caption{The MLFNN architecture with four dense layers.}
    \label{MLFNNfig}
    \end{figure}
    \item \textit{Bidirectional Encoder Representations from Transformers}: When working with Transfer learning and BERT, the focus is more on the fine-tuning aspect of BERT. A common way of doing this, is by adjusting the learning rate. The learning rates employed, were according to \cite{b20} and their research on BERT, whereby they employed learning rates: 5e-5; 4e-5; 3e-5; and 2e-5. The accuracy measurements, as well as the other previously mentioned performance metrics were then observed, before selecting the preferred model. Fig.~\ref{BERTmodel_plot} illustrates the BERT architecture with four dense layers. 
    \begin{figure}[htbp]
    \centerline{\includegraphics[scale=0.6]{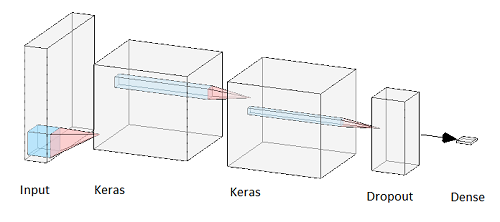}}
    \caption{BERT Model Architecture.}
    \label{BERTmodel_plot}
    \end{figure}
\end{enumerate}

\section{Results and Discussion}
This section presents the results obtained from the three models after varying model specific parameters in order to obtain the highest performance for a model on the aforementioned metrics during training, validation and testing. Discussions on how the best parameter settings were determined are also conducted. 

\subsection{Convolutinal Neural Network}
As previously mentioned, 3 CNN architectures were created, a single Conv1D/MaxPooling, a double Conv1D/MaxPooling, and a triple Conv1D/MaxPooling. These models were then trained and tested, and the performance was captured. Table ~\ref{CNNTab} shows training accuracy and loss along with validation accuracy and loss for three CNN architectures during training. Whereas, Table ~\ref{CNNTab2} shows precision, recall and F1-score for the three architectures during testing. From these results, there was not much difference amongst the CNN models. However, it was observed that the 2 Conv1D/MaxPooling layer architecture, performed slightly better in the accuracy and loss measurements, as well as in the precision score. Therefore, it was selected as the preferred model. 

\begin{table}[htbp]
\caption{CNN Model Comparison during Training}
\begin{center}
\begin{tabular}{ |p{2cm}|p{1cm}|p{1cm}|p{1cm}|p{1cm}| }
 \hline
 \multicolumn{5}{|c|}{CNN Model Accuracy and Loss} \\
 \hline
Architecture& TrainAcc & TrainLoss &ValAcc &ValLoss\\
 \hline
 1 Conv1D& 0.9899   &0.0507&0.9500& 0.2140\\
 2 Conv1D & \textbf{0.9966} & \textbf{0.0370}   &\textbf{0.9833}& \textbf{0.1800}\\
 3 Conv1D &0.9831 & 0.0690&  0.9167&0.3404\\
 \hline
\end{tabular}
\label{CNNTab}
\end{center}
\end{table}

\begin{table}[htbp]
\caption{CNN Model Comparison during Testing}
\begin{center}
\begin{tabular}{ |p{2cm}|p{1cm}|p{1cm}|p{1cm}| }
 \hline
 \multicolumn{4}{|c|}{CNN Model Precision, Recall and F1-Score} \\
 \hline
Architecture& Precision & Recall &F1-Score\\
 \hline
 1 Conv1D& 0.9545   &0.7241&0.8235\\
 2 Conv1D& \textbf{0.9565} & 0.7586 &0.8462\\
 3 Conv1D&0.9231 & \textbf{0.8276} & \textbf{0.8727}\\
 \hline
\end{tabular}
\label{CNNTab2}
\end{center}
\end{table}

\subsection{Multi-Layer Feed-Forward Neural Network}
As mentioned, when creating the model architecture, 3 models of varying number of dense layers(2 layer model, 3 layer model, 4 layer model) were created. Table ~\ref{MLFNNTab} displays the accuracy and loss measurements of the MLFNN models during training, whereas Table ~\ref{MLFNNTab2} displays the performance metrics of the three MLFNN architectures during testing. It is shown that the 2 Dense layer architecture, achieved the best scores in the precision, recall, and F1-score, but only by a slight difference, as well has having the same recall value as the 4 Dense layer architecture. However, the 4 Dense layer architecture, had the lowest loss values amongst the 3, as well as sharing the highest accuracy measurements with the 3 Dense layer architecture. Therefore, the 4 Dense layer architecture was selected as the preferred MLFNN model. The model layout is shown in Fig. ~\ref{MLFNNfig}.

\begin{table}[htbp]
\caption{MLFNN Model Comparison during Training}
\begin{center}
\begin{tabular}{ |p{2cm}|p{1cm}|p{1cm}|p{1cm}|p{1cm}| }
 \hline
 \multicolumn{5}{|c|}{CNN Model Accuracy and Loss} \\
 \hline
Architecture& TrainAcc & TrainLoss&ValAcc&ValLoss\\
 \hline
 2 Dense& 0.9865   &0.3207&0.9333& 0.4598\\
 3 Dense&\textbf{0.9899}  & 0.0692   &\textbf{0.9500}& 0.2303\\
 4 Dense&\textbf{0.9899} & \textbf{0.0397}&  \textbf{0.9500}&\textbf{0.1858}\\
 \hline
\end{tabular}
\label{MLFNNTab}
\end{center}
\end{table}

\begin{table}[htbp]
\caption{MLFNN Model Comparison during Testing}
\begin{center}
\begin{tabular}{ |p{2cm}|p{1cm}|p{1cm}|p{1cm}| }
 \hline
 \multicolumn{4}{|c|}{CNN Model Precision, Recall and F1-Score} \\
 \hline
 Architecture& Precision & Recall &F1-Score\\
 \hline
 \textbf{2 Dense}& \textbf{1.0}   &\textbf{0.7241}&\textbf{0.8400}\\
 3 Dense& 0.9524 & 0.6897 &0.8000\\
 4 Dense &0.9546 & \textbf{0.7241} & 0.8235\\
 \hline
\end{tabular}
\label{MLFNNTab2}
\end{center}
\end{table}
\subsection{Bidirectional Encoder Representations from Transformers}
Lastly, for the development of the BERT model, as mentioned the actual layout of the BERT model was not altered. However the fine-tuning method of adjusting the learning rates was employed. The employed learning rates of 2e-5, 3e-5, 4e-5, and 5e-5, in accordance with \cite{b20}. From this, the best performing model was selected as the preferred model.  Table ~\ref{BERTTab} presents the accuracy and loss measurements of the four BERT architectures during training. On the other hand, Table ~\ref{BERTTab2} presents the performance metrics of the four architectures during testing. As shown in these results, two of the learning rates, namely 3e-5, and 5e-5, returned 0.0 for Precision, Recall, and F1-score, and are therefore not able to be the preferred model. Amongst 2e-5 and 4e-5, 2e-5 obtained higher Precision, Recall, and F1-score, however achieved lower accuracy and loss values. Therefore, the model with the learning rate 2e-5 was selected as the preferred BERT model.

\begin{table}[htbp]
\caption{BERT Model Comparison during Training}
\begin{center}
\begin{tabular}{ |p{2cm}|p{1cm}|p{1cm}|p{1cm}|p{1cm}| }
 \hline
 \multicolumn{5}{|c|}{BERT Model Accuracy and Loss} \\
 \hline
Learning Rate& TrainAcc & TrainLoss &ValAcc&ValLoss\\
 \hline
 2e-5& 0.7399 & 0.6168 & 0.8500 & 0.5771\\
 3e-5& 0.7432 & 0.5515 & \textbf{0.9000} & 0.4742\\
 4e-5& \textbf{0.7466} & \textbf{0.5498} & \textbf{0.9000} & 0.4659\\
 5e-5& 0.7432 & 0.5544 & \textbf{0.9000}  & \textbf{0.4517}\\
 \hline
\end{tabular}
\label{BERTTab}
\end{center}
\end{table}

\begin{table}[htbp]
\caption{BERT Model Comparison during Testing}
\begin{center}
\begin{tabular}{ |p{2cm}|p{1cm}|p{1cm}|p{1cm}| }
 \hline
 \multicolumn{4}{|c|}{BERT Model Precision, Recall and F1-Score} \\
 \hline
 Learning Rate& Precision & Recall &F1-Score\\
 \hline
 2e-5& \textbf{0.2857} & \textbf{0.0699} &\textbf{0.1111}\\
 3e-5& 0.0    & 0.0    & 0.0\\
 4e-5& 0.1667 & 0.0345 & 0.0571\\
 5e-5& 0.0      & 0.0     & 0.0\\
 \hline
\end{tabular}
\label{BERTTab2}
\end{center}
\end{table}

\subsection{Discussion}
Now with the three preferred models from each type, the selection of the final model is required. These performances are collated in Table~\ref{ModelTab} and Table~\ref{ModelTab2}. From these results, the CNN model performed performs slightly better than the MLFNN. The BERT model however, proved to not only be inefficient, as the training time taken was much longer than the CNN and the MLFNN, it also provided undesirable results, as shown by the performance metrics. This lines up similarly with \cite{b21}, whereby they compared BERT with Learning and Teaching Support Material (LSTM) on a small dataset, and found the LTSM to perform better than the BERT model. From everything discussed, it is concluded that the CNN is the preferred model for the task in question and was therefore employed as the text classification model, to assist in the obtainment of hijacking incident reports from the tweets database.

\begin{table}[htbp]
\caption{Model Comparison}
\begin{center}
\begin{tabular}{ |p{2cm}|p{1cm}|p{1cm}|p{1cm}|p{1cm}| }
 \hline
 \multicolumn{5}{|c|}{Model Accuracy and Loss} \\
 \hline
Architecture& TrainAcc &TrainLoss&ValAcc&ValLoss\\
 \hline
 CNN    &\textbf{0.9966}&\textbf{0.0370}&\textbf{0.9833}&\textbf{0.1800}\\
 MLFNN  &0.9899&0.0397&0.9500&0.1858\\
 BERT   &0.7399&0.6168&0.8500&0.5771\\
 \hline
\end{tabular}
\label{ModelTab}
\end{center}
\end{table}

\begin{table}[htbp]
\caption{Model Comparison 2}
\begin{center}
\begin{tabular}{ |p{2cm}|p{1cm}|p{1cm}|p{1cm}| }
 \hline
 \multicolumn{4}{|c|}{ Model Precision, Recall and F1-Score} \\
 \hline
 Architecture& Precision & Recall &F1-Score\\
 \hline
 CNN    &\textbf{0.9565}&\textbf{0.7586}&\textbf{0.8462}\\
 MLFNN  &0.9546&0.7241& 0.8235\\
 BERT   &0.2857&0.0699& 0.1111\\
 \hline
\end{tabular}
\label{ModelTab2}
\end{center}
\end{table}

\subsection{Vehicle Hijacking Maps}
\begin{figure}[htbp]
\centerline{\includegraphics[scale=1]{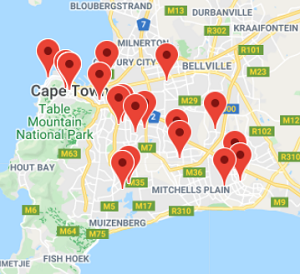}}
\caption{Cape Town Point-map for hijacking spots}
\label{CapeTownPointmap}
\end{figure}

As mentioned, the goal of devising and obtaining a machine learning model was to determine relevant tweets relating to hijackings in the Cape Town area. The tweets that were deemed relevant by the CNN model were processed further in order to extract the locations mentioned in the text in order to plot them on the maps. In order to obtain the locations, the Geopy \cite{b22} python library was used. Next, Geograpy \cite{b23} was used to obtain the coordinates of these locations. This information was used to create a heat-map, as well as a point-map by employing the Gmaps \cite{b24} library. The point-map of hijacking spots, in Fig.~\ref{CapeTownPointmap}, displays the locations mentioned in the CNN identified relevant tweets.
%

\section{Conclusion}
This work was aimed at devising a deep learning learning model for text classification of relevant tweets containing the keyword "hijacking". The goal was to use Twitter as a data source of hijacking reports to display the incident locations on maps for visualisation. Three deep learning models, were therefore trained on 296 tweets and tested on 130 tweets. These models are as follows: a Convolutional Neural Network (CNN), a Multi-Layer Feed-Forward Neural Network (MLFNN), and a Bidirectional Encoder Representations from Transformers (BERT) model. The CNN and MLFNN were tested with varying architectures, and the BERT model was tested with varying learning rates. Through testing and observation, it was found that the CNN model was the highest performing model. It was, therefore, employed to identify relevant tweets from a database of tweets containing the keyword "hijacking". Through this, the relevant hijacking tweets were identified, and the in-text mentioned locations of the incidents were extracted. This allowed for the creation of the point-map of hijacking spots in the Cape Town area.

In the future, the use of evolutionary computation will be explored with the aim of optimizing the deep learning models. Furthermore, a mobile application is under development to make this information usable by the general public.

\end{document}